\pdfoutput=1

\documentclass[11pt]{article}

\usepackage[final]{acl}

\usepackage{times}
\usepackage{latexsym}
\usepackage{multirow}
\usepackage{amsmath}
\usepackage{booktabs}
\usepackage{xcolor}
\usepackage{tabularx}
\usepackage{graphicx}
\usepackage{algorithm}
\usepackage{algorithmic}
\usepackage{changes}
\usepackage[T1]{fontenc}

\usepackage[utf8]{inputenc}

\usepackage{microtype}

\usepackage{inconsolata}
\usepackage{tcolorbox}

\usepackage{array}
\usepackage{lipsum} 
\usepackage{arydshln}
\usepackage{float}
\definecolor{lightgray}{gray}{0.9}
\definecolor{lightgray}{gray}{0.9}

\title{Streamlining the Collaborative Chain of Models \\ into A Single Forward Pass in Generation-Based Tasks}

\newcommand*{\affaddr}[1]{#1}

\newcommand*{\email}[1]{\texttt{#1}}

\author{Yuanjie Lyu, Chao Zhang,  Yuhao Chen, Yong Chen, Tong Xu\thanks{Corresponding author.}\\
\affaddr{University of Science and Technology of China} \\
\email{\{s1583050085\}@gmail.com,}\\
\email{\{tongxu\}@ustc.edu.cn} \\
}

\begin{document}
\maketitle
\renewcommand{\thefootnote}{\fnsymbol{footnote}}
\renewcommand{\thefootnote}{\arabic{footnote}}
\begin{abstract}
In Retrieval-Augmented Generation (RAG) and agent-based frameworks, the "Chain of Models" approach is widely used, where multiple specialized models work sequentially on distinct sub-tasks. This approach is effective but increases resource demands as each model must be deployed separately. Recent advancements attempt to address this by applying prompt tuning, which allows a shared base model to adapt to multiple tasks with minimal parameter changes. However, a key challenge remains: intermediate outputs, passed between models as plain text, require recomputation of hidden states (i.e., Key and Value (KV) states in Transformers) during inference. In this paper, we introduce FTHSS, a novel prompt-tuning method that enables models to share KV hidden states, eliminating redundant forward passes and reducing KV cache storage. By modifying input and attention masks during training, FTHSS allows models to effectively utilize KV hidden states from prior models in both single- and multi-round scenarios. Empirical results on four tasks show that FTHSS matches the performance of traditional model chains while improving inference efficiency.
\footnote{Code: 
\href{https://github.com/haruhi-sudo/FTHSS}{https://github.com/haruhi-sudo/FTHSS}.
}
\end{abstract}

\section{Introduction}
\begin{figure}[t] 
\centerline{
\includegraphics[width=0.5\textwidth]{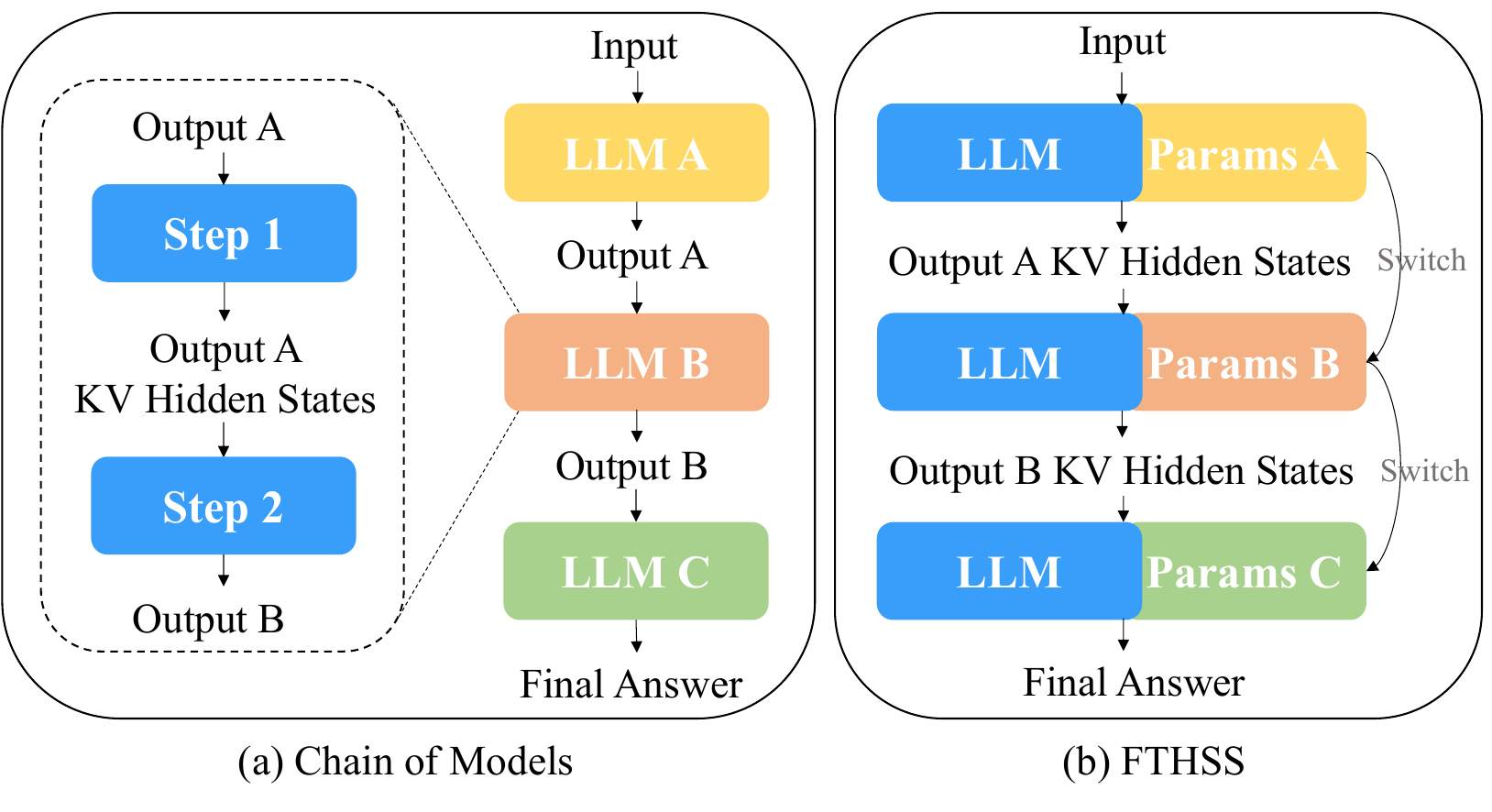}} 
\caption{ 
Comparison of "Chain of Models" (a) and FTHSS (b): In (a), models sequentially pass outputs as plain text, requiring KV recomputation. In (b), FTHSS shares KV hidden states, reducing redundant forward passes. PEFT methods allow the deployment of multiple models on a single device, with parameters changing, so there is no communication overhead for hidden states.
} 
\label{fig:demo} 
\end{figure}

In many Retrieval-Augmented Generation (RAG) and agent-based frameworks~\cite{lewis2020retrieval}, multiple Large Language Models (LLMs) often collaborate sequentially. Each model focuses on a specific sub-task and passes its output as input to the next model until the task is completed\cite{zhang2024chain}. For instance, some RAG post-retrieval optimization methods~\cite{xu2023recomp,kim2024sure} involve summarizing retrieved documents with a summarization model, and then generating answers with a question-answering model. 
These stepwise approaches leverage the strengths of individual models and have proven effective in many scenarios. As a result, the "Chain of Models" approach has gained popularity~\cite{zhang2024chain}.

Deploying every specialized LLM in such chains significantly increases the resources needed. To address this, researchers have explored parameter-efficient fine-tuning (PEFT) methods, such as prompt tuning~\cite{liu2021p} and LoRA~\cite{hu2021lora}. These techniques allow fine-tuning with a fraction of the parameters when training. During inference, a shared base model is deployed on a single device and handles multiple tasks with distinct parameter configurations. This approach merges "Chain of Models" workflows into a single architecture, adapting to various sub-tasks through selective parameter usage.
However, a critical bottleneck remains: in the chain, the intermediate key-value(KV) hidden states from one model cannot be directly reused by the next model due to parameter differences. As a result, communication between models in the chain relies on passing plain text, forcing the downstream model to recompute hidden states. This practice not only adds computational overhead, but also raises KV cache storage requirements for each model in the chain, further hampering efficiency.

In this paper, we argue that such recomputation is unnecessary. Even with parameter differences, the KV hidden states produced by one model should only differ marginally from those recalculated by the next. Particularly in prompt-tuning methods, the KV hidden states produced by the previous model are essentially conditioned on a few noisy tokens.
With appropriate fine-tuning, the subsequent model can effectively interpret and utilize the KV hidden states of the previous model despite these noises, as Figure~\ref{fig:demo} shows. 

To realize this vision, we propose FTHSS(Fine-Tuning for Hidden State Sharing), a prompt-tuning-based method that enables models in a chain to share KV hidden states. Specifically, when fine-tuning the model in single-round scenarios, where each model is invoked only once, we use KV hidden states from the prior models as input rather than plain text. This training approach requires extensive storage and access to KV hidden states, which may potentially increase training time and storage demands. To mitigate this, we introduce an online optimization strategy. By modifying the input and attention mask for each layer, we recompute the prior model's KV hidden states in memory during training, thus avoiding the overhead of storage and access. In multi-round scenarios, where models in the chain are invoked repeatedly, each model must adapt to the KV hidden states of others, so all models in the chain are trained synchronously to ensure mutual adaptation. 
After fine-tuning, models can dynamically switch learnable prompt tokens during inference, adapting based on task requirements, while leveraging precomputed KV cache for direct generation. And since prompt-tuning-based methods enable the deployment of multiple models on a single device, communication overhead for hidden states is effectively eliminated.


Empirical results on four tasks, including single-round and multi-round, demonstrate that FTHSS leads to a comparable performance to the chain of models, 
while enhancing inference efficiency.
Technical contributions of this paper can be summarized as follows:
\begin{enumerate}
\item[\textbullet]To the best of our knowledge, we are the first to streamline the chain of models by sharing KV hidden states, thereby reducing the need for recomputing intermediate results.

\item[\textbullet]We introduce a prompt-tuning-based training strategy, FTHSS, that supports KV hidden state sharing across models in both single-round and multi-round scenarios.

\item[\textbullet] Experimental results show that FTHSS maintains comparable performance while significantly reducing inference latency and eliminating redundant KV cache storage.
\end{enumerate}

\section{Related Work}
\subsection{Chain of Models}

The Chain of Models approach sequentially links specialized models, using the output of one as the input for the next~\cite{zhang2024chain}. This method allows for incremental processing of sub-tasks, and has been widely adopted across various domains. For example, Retrieval-Augmented Generation (RAG)~\cite{lewis2020retrieval} improves the performance of question-answering (QA) tasks by combining retrieval and generation models. Additionally, the Chain of Models framework has proven highly effective for mathematical reasoning\cite{sun2023corex, dong2024effiqa, lei2024macm} and long-text generation~\cite{xi2025omnithink, wang2024autopatentmultiagentframeworkautomatic}.


While leveraging specialized models improves performance, it also increases deployment costs. One optimization strategy is to consolidate multiple models into a single, unified model through distillation. For instance, GritLM~\cite{muennighoff2024generative} enables task-switching through instruction modifications, combining retrieval and generation. OneGen~\cite{zhang2024onegen} introduces retrieval tokens, allowing LLMs to handle both tasks in a single forward pass. RankRAG~\cite{yu2024rankrag} integrates ranking and generation into a single retrained model. However, these methods require the distilled model to perform well in multiple tasks, which remains a significant challenge. The FTHSS method proposed in this paper diverges from the distillation paradigm, and it still leverages the strengths of multiple models while reducing the demand for computing resources.

\subsection{KV Cache Compression and Sharing}

Large Language Models (LLMs) face significant bottlenecks due to high memory and computational demands, with the key-value (KV) cache being a major contributor. The KV cache stores the keys and values for each Transformer layer during generation to avoid redundant computations. During deployment, the KV cache can occupy over 30\% of GPU memory~\cite{kwon2023efficient}. 

Some straightforward approaches address this issue by compressing context length~\cite{ge2024incontextautoencodercontextcompression, jiang2023llmlingua,li2023compressing} or employing sparse attention matrices ~\cite{xiao2023efficient, han2023lm}. More recently, methods focusing on KV cache reuse have been proposed. YOCO~\cite{sun2024you} utilizes a cross-decoder mechanism with cross-attention to reuse cached values, allowing the model to store KV pairs only once while maintaining global attention capabilities. LCKV~\cite{wu2024layer} and KVSharer~\cite{yang2024kvsharerefficientinferencelayerwise} enable KV cache sharing across layers within the same model. While these methods effectively enhance model efficiency by reusing and sharing KV caches at different layers of a single model, FTHSS extends this concept to multiple models.

\section{Methodology}




\begin{figure*}[t] 
\centerline{
\includegraphics[width=1.\textwidth]{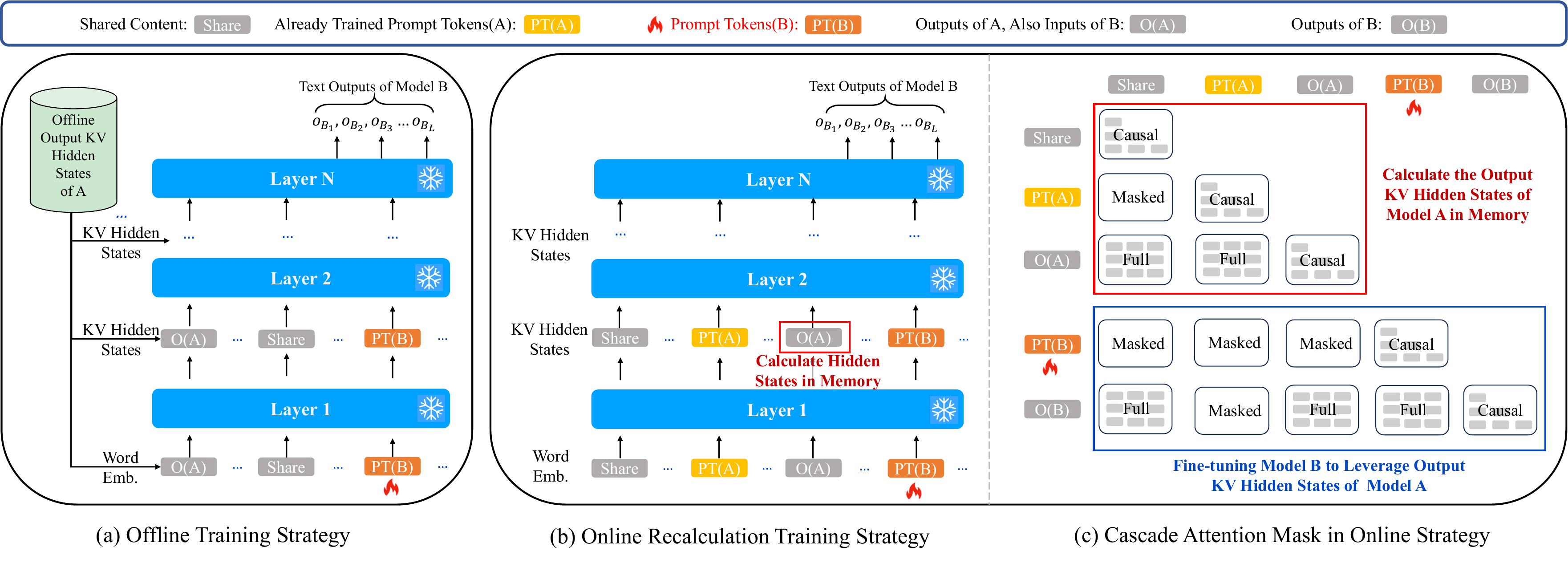}} 
\caption{ 
An example of fine-tuning model B in the model chain A → B. For simplicity, the unique inputs of model A and model B are omitted. \textbf{Left:} Offline fine-tuning, where the output KV hidden states of fully trained model A are stored and used as input for model B. \textbf{Middle:} Online, where the output KV hidden states of model A are recalculated in memory. \textbf{Right:} We calculate the output KV hidden states of model A in memory and fine-tune model B by adjusting the attention mask for each layer. We use the online training strategy in practical applications.
} \label{fig:single-turn} 
\end{figure*}

In this section, 
we begin by highlighting a key challenge: model chains rely on text-based communication, which prevents the direct transfer of KV hidden states between models. We then explore the feasibility of fine-tuning the downstream model to process KV hidden states from the upstream model, although these hidden states often include noise tokens irrelevant to the downstream task.
Lastly, we propose training strategies, FTHSS, to achieve KV hidden state sharing. 


\subsection{Preliminary}
Multiple models $ M_1, M_2, \ldots, M_n $ often collaborate sequentially in RAG and agent-based tasks, with each model $ M_i $ handling a specific task component. Specifically, model $M_i$ processes the output $T_{i-1}$ from the previous model, along with its unique input $x_i$, to produce output $T_i$ for the next model. This process is expressed as: $T_i = M_i(T_{i-1}, x_i)$, forming a chain of models.

Given the high cost of deploying all models in such a chain, we can adopt a prompt-tuning approach. A shared base model $M_{\theta}$ is fine-tuned to perform different tasks, with each model $M_i$ distinguished solely by its fine-tuned prompt tokens $P_i$. This approach allows us to deploy only $M_{\theta}$, dynamically adjusting prompt tokens to replicate the behavior of multiple models:
\begin{equation}
T_i = M_{\theta}(T_{i-1}, x_i, P_i).
\end{equation}

While this approach simplifies the model chain, communication between models still occurs via text. Upon receiving the output  $T_{i-1}$ from the previous model, each model $M_i$ recalculates the hidden state of $T_{i-1}$ based on its prefix $P_i$(a process known as "prefilling"), and then generates the output $T_i$ and the corresponding hidden states $O_{T_{i}}$ autoregressively(a process known as "decoding"):
\begin{equation}
H_{T_{i-1}}, H_{x_{i}}, H_{P_{i}} = \text{Prefilling}(T_{i-1}, x_i, P_i),
\end{equation}
\begin{equation}
T_i, O_{T_{i}} = \text{Decoding}(H_{T_{i-1}}, H_{x_{i}}, H_{P_{i}}),
\end{equation}
where $T_i$ is the output text and $O_{T_{i}}$ is the output hidden states of $T_i$.

In this paper, we argue that recalculating the KV hidden state $H_{T_{i-1}}$ is unnecessary. Instead, model $M_i$ can directly use KV hidden states $O_{T_{i-1}}$ output by the previous model $M_{i-1}$ as inputs. Besides, since prompt tuning allows the deployment of multiple models on a single device, there is no communication overhead of hidden states.




\subsection{Fine-Tuning for Hidden State Sharing}
\label{sec:fine-tuning}
Based on the above analysis, we aim to ensure that the KV hidden states computed by the previous model can be directly interpreted by the next. 
This is feasible due to the minimal differences between $H_{T_{i}}$ and $O_{T_{i}}$. Since models fine-tuned with prompt tuning on the same base model share identical structures and parameters, they differ only in the fine-tuned prompt tokens and input data.

Specifically, the output KV hidden state of $M_{i}$ during generation of the $j+1$-th token:
\begin{equation}
\_, O_{T_{i,j}} = \text{Decoding}(T_{i, 1:j}, H_{T_{i-1}}, H_{P_{i}}),
\label{eq:h_1}
\end{equation}
where $O_{T_{i,j}}$ is the hidden state of token $T_{i,j}$ output by $M_i$. We ignore the unique input for simplicity. 

When $T_{i,j}$ serves as the input of $M_{i+1}$ rather than the output of $M_{i}$, the KV hidden state must be recomputed:
\begin{equation}
H_{T_{i,j}} = \text{Prefilling}(T_{i, 1:j}),
\label{eq:h_2}
\end{equation}
where $H_{T_{i,j}}$ is the KV hidden state of token $T_{i,j}$ calculated by $M_{i+1}$. We omit the prefix $P_{i+1}$ as it can be appended after $T_{i,1:j-1}$.

Since the attention calculation method is the same in both prefilling and decoding stages, the difference between equations (\ref{eq:h_1}) and (\ref{eq:h_2}) is minimal, with only the prefixes and inputs differing. This suggests that the output hidden state of $M_{i}$ introduces minimal noise for $M_{i+1}$,
and fine-tuning may be a practical solution.

We propose FTHSS (Fine-Tuning for Hidden State Sharing), a fine-tuning method to minimize these differences. By fine-tuning model $M_i$ with noisy KV hidden states from model $M_{i-1}$ as input, rather than the original ones, performance can be maintained despite the noise. We are currently exploring the implementation of this process.

\subsubsection{Fine-Tuning Strategies of Single-Round}
In practical applications, model chains are deployed in two configurations: single-round, where each model is called once, and multi-round, where models may be invoked multiple times. These configurations require distinct fine-tuning strategies.

Consider a model chain consisting of  A and B in a single-round scenario, where model A precedes model B, and its output serves as B's input. The training data and processes are organized as:



\paragraph{Model Input}
Since model A is the first in the chain, it does not require adjustment to any preceding model's input. Thus, the fine-tuning data for model A follows standard prompt tuning. However, we refine this process by reordering the input:
\begin{itemize}
\item Model A input order: shared content tokens, learnable prompt tokens (A), unique input content tokens for A.
\end{itemize}

We place the shared content before the learnable prompt tokens. Since the shared content is used across all models in the chain, this arrangement ensures that the KV hidden states of the shared content remain unaffected by the learnable tokens, thereby preventing the introduction of noise.

Since the output of model A serves as the input for model B,  A must be fully fine-tuned before fine-tuning B. Besides, the input to model B should consist of the output KV hidden states from A, rather than the tokens generated by A.

\begin{itemize} 
\item Model B input order: shared content tokens, output KV hidden states of fine-tuned model A, learnable prompt tokens (B), and unique content tokens for B.
\end{itemize}

\paragraph{Fine-Tuning Process}
As mentioned earlier, model B must be trained after model A, using the output KV hidden states from A. The fine-tuning process for the model chain proceeds as follows:

\begin{itemize}
\item Fine-tune A to generate output A. 
\item Store the output KV hidden states from the fully fine-tuned model A. 
\item Offline load the hidden states and fine-tune B to leverage them in generating output B. 
\end{itemize}

When fine-tuning model B, the position ID should not start at 0. Since model A's hidden states already contain position information, the position IDs for model B should begin at $l+1$, where $l$ is the last position ID in model A. As the LLM in this paper employs relative position encoding (e.g., RoPE~\cite{su2024roformer}), the absolute position is not critical. Therefore, the position ID ranges $[0, 1, \dots, l]$  and  $[l+1, l+2, \dots, 2l+1]$  are equivalent for attention computation. The proof is provided in Appendix \ref{sec:position}.

\paragraph{Fine-Tuning Tricks to Save Storage}
Given that most existing LLMs are based on the Transformer architecture, they typically include numerous layers and attention heads. As Figure \ref{fig:single-turn}(a) shows, the approach described above requires storing and accessing a large number of KV hidden states, which can be impractical. To address this, we propose recomputing the output KV hidden states of model A in memory, rather than storing them offline, as illustrated in Figure \ref{fig:single-turn}(b). 

Specifically, during the training of model B, we modify the input to B as follows:
\begin{itemize} 
\item Model B input order: shared content tokens, fine-tuned prompt tokens (A), unique input content tokens for A, output tokens of A, learnable prompt tokens (B), and unique content tokens for B. 
\end{itemize}

Notably, We incorporate the fine-tuned prompt tokens (A),  along with both the input and output tokens of model A, as part of model B's input. By adjusting the attention mask, we calculate model A's output KV hidden states in memory (red box in Figure \ref{fig:single-turn}(c)). Simultaneously, the learnable prompt tokens (B) are fine-tuned to generate model B’s output, using the recalculated KV hidden states from model A (blue box in Figure \ref{fig:single-turn}(c)).

The above algorithm outlines the fine-tuning process for a simplified model chain A$\rightarrow$B. In practical applications, when more than two models are involved in a model chain, each model can be trained sequentially, following the order of the chain. During this process, each model's input and attention mask should be adjusted accordingly.

\begin{figure}[t]
\centerline{\includegraphics[width=0.46\textwidth]{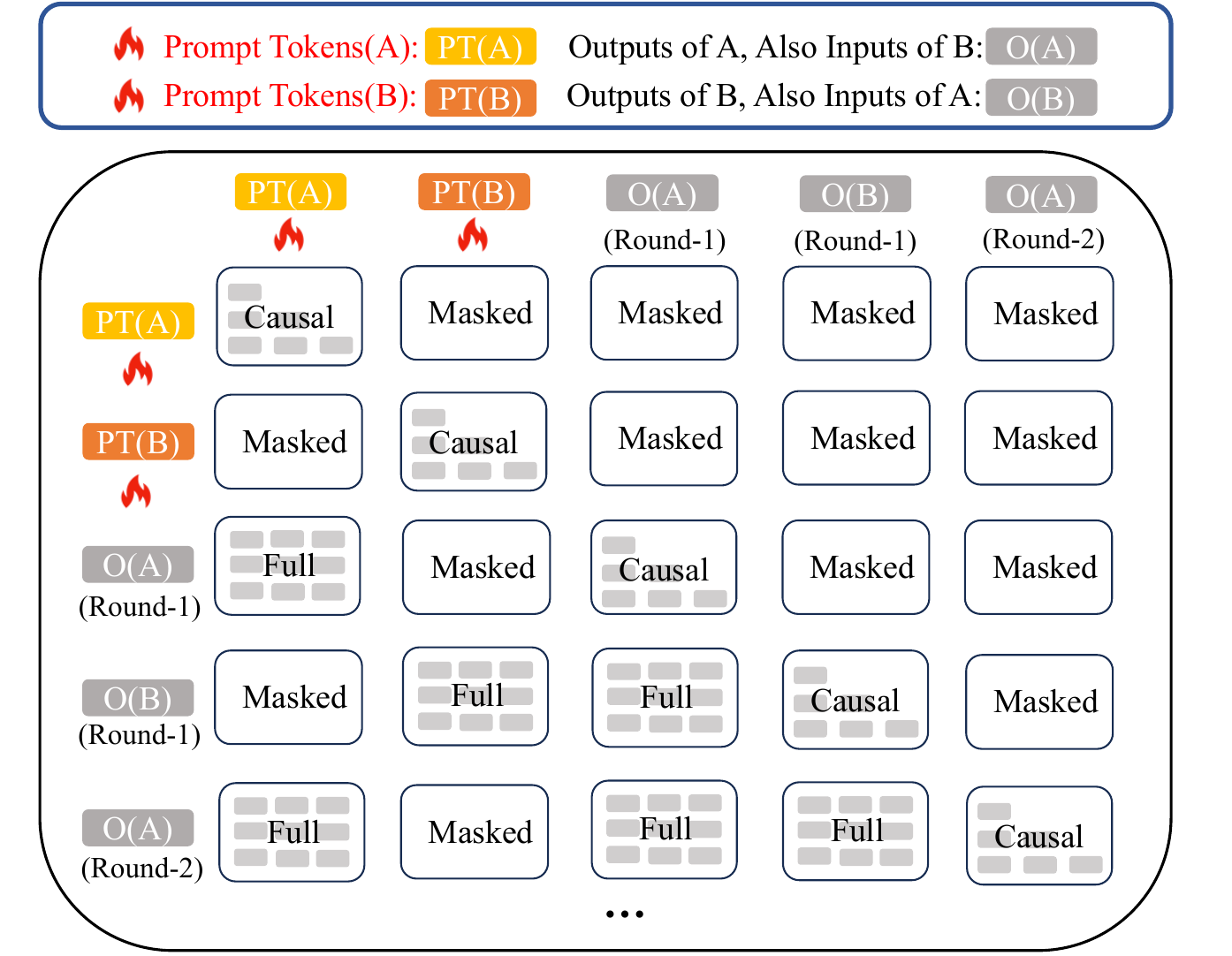}}
\caption{Cascade attention mask for every layer in the multi-round scenario.
}
\label{fig:multi-turn}
\end{figure}

\subsubsection{Fine-Tuning Strategies of Multi-Round}
In a multi-round scenario, models may be invoked sequentially multiple times, allowing for more complex chains, such as $A \rightarrow B \rightarrow A \rightarrow B$. In this context, model B must adapt to the output of model A, while model A must also adapt to the output of model B. This differs from a single-round scenario, since models must be fine-tuned simultaneously.

To address this challenge, we modify the inputs and attention masks for both models, as illustrated in Figure~\ref{fig:multi-turn}. Specifically, the prompt tokens for both models are positioned at the beginning of the input. When computing the loss on the output of model A, attention scores are computed while masking the prompt tokens of model B. Conversely, When computing the loss on the output of model B, the prompt tokens of model A are masked. This ensures that model A's tasks are guided solely by its own prompt tokens, while model B's tasks are directed by its respective prompt tokens.

\section{Experiments}
\begin{table*}[t]
    \centering
    \begin{minipage}{0.5\textwidth}
        \centering
        \resizebox{\textwidth}{!}{
        \begin{tabular}{ccccccc}
            \toprule
            \textbf{Task (\(\rightarrow\))} & \multicolumn{6}{c} {Context Compression \& QA} \\

            \textbf{Dataset (\(\rightarrow\))} & \multicolumn{2}{c}{HQA} & \multicolumn{2}{c}{TQA} & \multicolumn{2}{c}{NQ} \\

            \textbf{Metric (\(\rightarrow\))} & EM & F1 & EM & F1 & EM & F1  \\
            \midrule

            \multicolumn{7}{c}{\textbf{Single Model}} \\
            Native & 14.4 & 22.8 & 40.1 & 53.7 & 14.5 & 26.4\\
            Standard RAG & 24.0 & 36.2 & 47.0 & 58.3 & 28.5 & 44.8 \\
            Prompt Tuning & 26.0 & 36.2 &26.4 & 44.2 & 32.7 & 45.1 \\

            \midrule
            \multicolumn{7}{c}{\textbf{Chain of Models}} \\
            Compress\&QA & \textcolor{gray}{\textbf{30.4}} & \textcolor{gray}{\textbf{43.8}} & \textcolor{gray}{\textbf{59.7}} & \textcolor{gray}{\textbf{68.3}} & \textcolor{gray}{\textbf{35.0}} & \textcolor{gray}{\textbf{48.3}} \\        
            \midrule  

            \multicolumn{7}{c}{\textbf{Streamlining}} \\
            Distill & 28.3 & 42.1 & 54.3 & 63.9 & 21.4 & 33.1 \\
            FTHSS(Our) & \textbf{29.0} & \textbf{42.2}& \textbf{59.3} & \textbf{67.5} & \textbf{35.8} & \textbf{45.6} \\
            \bottomrule
        \end{tabular}
        }
        \caption{
        Performance on the single-round task: Compression\&QA for FTHSS and other methods. \textbf{Bold numbers} indicate the best performance, except for the original chain of models (denoted in gray). Same below.
        }
        \label{tab:recomp_performance}
    \end{minipage}%
    \hfill
    \begin{minipage}{0.465\textwidth}
        \centering
        \resizebox{\textwidth}{!}{
        \begin{tabular}{ccccccc}
            \toprule
            \textbf{Task (\(\rightarrow\))} & \multicolumn{4}{c} {Query Rewriting \& QA} \\

            \textbf{Dataset (\(\rightarrow\))} & \multicolumn{2}{c}{HQABM25} & \multicolumn{2}{c}{2WikiBM25} \\

            \textbf{Metric (\(\rightarrow\))} & EM & F1 & EM & F1 \\
            \midrule

            \multicolumn{5}{c}{\textbf{Single Model}} \\
            Native  & 13.4 & 19.5 & 13.8 & 21.4 \\
            Standard RAG  & 19.0 & 31.1 & 14.4 & 21.6\\
            Prompt Tuning  & 18.2 & 29.8 & 20.6 & 27.4 \\

            \midrule
            \multicolumn{5}{c}{\textbf{Chain of Models}} \\
            Rewrite\&QA & \textcolor{gray}{\textbf{27.0}} & \textcolor{gray}{\textbf{37.2}} & \textcolor{gray}{\textbf{24.4}} & \textcolor{gray}{\textbf{30.2}} \\        
            \midrule 

            \multicolumn{5}{c}{\textbf{Streamlining}} \\
            Distill  & 20.8 & 30.4 & 18.0 & 23.9 \\
            FTHSS(Our)  & \textbf{27.4} & \textbf{36.6}& \textbf{24.0} & \textbf{29.9} \\
            \bottomrule
        \end{tabular}
        }
        \caption{
            Performance on the single-round task: Query Rewrite\&QA for FTHSS and other methods.
        }
        \label{tab:rewrite_performance}
    \end{minipage}
\end{table*}


\subsection{Setup}

We conduct experiments on both single-round and multi-round tasks. These experiments aim to evaluate whether the FTHSS approach can retain the functionality of model chains while improving inference efficiency in various scenarios.

\subsubsection{Single-Round Evaluation}
\paragraph{Tasks}
Many RAG frameworks involve chains of models due to their modular nature, 
making them suitable for our evaluation. Common RAG optimization methods include pre-retrieval and post-retrieval optimization. We select two tasks from each of them as benchmarks:
\begin{itemize}
\item Context Compression \& Question Answering
\item Query Rewriting \& Question Answering
\end{itemize}

The Context Compression \& QA task involves compressing retrieved content into a noise-free context for the final response. The Query Rewriting \& QA task rewrites the query to retrieve more relevant information, and then generates the final response.

For the training data of Context Compression \& QA task, we follow the data specified in ReComp\cite{xu2023recomp}, while for the training data of  Query Rewriting \& QA task, we adhered to the data outlined by ~\citet{ma2023query}.

\paragraph{Baselines}
In the experiment, we compare three types of methods: (1) direct answer from a single model (Native, Standard RAG, Prompt Tuning), (2) using a model chain to generate intermediate results, which are then used to provide the final answer (Compress\&QA, Rewrite\&QA), and (3) simplifying the model chain to perform similarly to a single model (Distill, FTHSS).
Distill refers to fine-tuning one model to generate all intermediate steps, effectively distilling the capabilities of multiple models into a single model. We use Llama-3-8B~\cite{dubey2024llama} as the base model for all models in the chain. To ensure a fair comparison, all fine-tuning techniques discussed in this paper employ prompt tuning~\cite{liu2021p}.

\paragraph{Datasets}
We use the following widely adopted datasets to validate our approach: Natural Questions (NQ)\cite{kwiatkowski2019natural}, TriviaQA (TQA)\cite{joshi2017triviaqa}, 2WikiMultiHopQA(2Wiki)~\cite{ho2020constructing} and HotpotQA (HQA)\cite{yang2018hotpotqa}. 

\subsubsection{Multi-Round Evaluation}
\paragraph{Tasks}

In multi-round scenarios, models in a chain are invoked repeatedly. We selected "Reasoning \& Memory" as a validation task~\cite{jin2024disentangling}, which decomposes the inference process into two iterative steps: (1) memory recall, retrieving relevant knowledge from the model's memory, and (2) reasoning, applying logical operations to the recalled knowledge. Additionally, we evaluate our methods on an active retrieval augmented generation task~\cite{jiang2023active}. The Active RAG task involves multiple rounds of retrieval, which actively decides what to retrieve across the course of the generation.

For the training data of Memory\&Reasoning task, we use the data from~\citet{jin2024disentangling}, while for the training data of Active RAG task, we follow the data proposed by \citet{lyu2024retrieve}.


\paragraph{Baselines}
The multi-round baselines are essentially identical to the single-round approach. They are categorized into three types: (1) direct answering (Single Model), (2) using a model chain to generate intermediate results (Memory\&Reason, Plan\&Generation), and (3) simplifying the model chain (Distill, FTHSS). 

\paragraph{Datasets}
We take the following widely adopted datasets for evaluation: StrategyQA~\cite{geva2021did}, TruthfulQA(TruthQA)~\cite{lin2021truthfulqa}, CommonsenseQA(ComQA)~\cite{talmor2018commonsenseqa}, PubHealth~\cite{zhang2023interpretable}, 2WikiMultiHopQA(2Wiki)~\cite{ho2020constructing} and HotpotQA (HQA)\cite{yang2018hotpotqa}. 

For more details, we explain each task and other hyper-parameters in the Appendix ~\ref{sec:hyper} and ~\ref{sec:task}.

\begin{table*}[t]
    \centering
    \begin{minipage}{0.48\textwidth}
        \centering
        \resizebox{\textwidth}{!}{
        \begin{tabular}{cccc}
            \toprule
            \textbf{Task (\(\rightarrow\))} & \multicolumn{3}{c} {Memory \& Reasoning} \\

            \textbf{Dataset (\(\rightarrow\))} & StrategyQA & ComQA & TruthQA \\
            \textbf{Metric (\(\rightarrow\))} & Acc & Acc & Acc   \\
            \midrule

            \multicolumn{4}{c}{\textbf{Single Model}} \\
            Zero-shot &  63.0	& 57.9 & 39.0\\ 
            CoT  & 63.0 &  66.1 & 47.6 \\
            Prompt Tuning & 63.6 & 66.7  & 65.2 \\

            \midrule
            \multicolumn{4}{c}{\textbf{Chain of Models}} \\
            Memory\&Reason & \textcolor{gray}{\textbf{70.1}} & \textcolor{gray}{\textbf{71.3}} & \textcolor{gray}{\textbf{69.2}} \\      
            \midrule 

            \multicolumn{4}{c}{\textbf{Streamlining}} \\
            Distill & 65.1 & 62.3 & 65.2 \\ 
            FTHSS(Our) & \textbf{69.2} & \textbf{70.3} & \textbf{68.9} \\
            \bottomrule
        \end{tabular}
        }
        \caption{
        Performance on the multi-round task: Memory \& Reasoning for FTHSS and other methods. 
        We evaluate the performance on multiple-choice questions using accuracy as the metric.
        }
        \label{tab:memory_performance}
    \end{minipage}%
    \hfill
    \begin{minipage}{0.47\textwidth}
        \centering
        \resizebox{\textwidth}{!}{
        \begin{tabular}{ccc}
            \toprule
            \textbf{Task (\(\rightarrow\))} & \multicolumn{2}{c} {Active RAG} \\

            \textbf{Dataset (\(\rightarrow\))} & Pubhealth & 2WikiBM25 \\
            \textbf{Metric (\(\rightarrow\))} & Acc & F1   \\
            \midrule

            \multicolumn{3}{c}{\textbf{Single Model}} \\
            Native &  69.5 & 21.4  \\ 
            Standard RAG & 56.1 & 21.6 \\
            Prompt Tuning & 69.1 & 27.4 \\

            \midrule
            \multicolumn{3}{c}{\textbf{Chain of Models}} \\
            Plan\&Generation & \textcolor{gray}{\textbf{73.4}} & \textcolor{gray}{\textbf{33.6}} \\        
            \midrule 

            \multicolumn{3}{c}{\textbf{Streamlining}} \\
            Distill & 70.1 & 23.1 \\ 
            FTHSS(Our) & \textbf{72.0}& \textbf{31.9} \\
            \bottomrule
        \end{tabular}
        }
        \caption{
        Performance on the multi-round task: Plan \& Retrieval for FTHSS and other methods.
        }
        \label{tab:rpg_performance}
    \end{minipage}
\end{table*}

\subsection{Main Results}
\paragraph{FTHSS leads to a comparable performance with the chain of models in both single-round and multi-round scenarios.}
We benchmark FTHSS with other models in Table \ref{tab:recomp_performance} and \ref{tab:rewrite_performance} in single-round settings, and find that FTHSS outperforms all single models while achieving comparable performance  to the chain of models. This demonstrates that our method avoids repeated computation of intermediate KV hidden states, improving efficiency without sacrificing performance.

For instance, in the Context Compression\&QA task on the TQA dataset, FTHSS achieves an EM score just 0.4 points lower than the approach using separate models, demonstrating nearly identical performance. Importantly, the compressed context no longer requires a forward pass through the QA model. Instead, it directly leverages the KV hidden states output by the compression model, reducing redundant computations and inference time.

Table \ref{tab:memory_performance} and \ref{tab:rpg_performance} present the results of multi-round experiments, which align closely with the findings from single-round experiments. This consistency highlights that, in addition to eliminating redundant intermediate computations, our method also removes the necessity of storing KV caches for individual models within the chain.

\paragraph{The chain of models outperforms single models.}

As shown in Tables \ref{tab:recomp_performance}, \ref{tab:rewrite_performance}, \ref{tab:memory_performance}, and \ref{tab:rpg_performance}, methods like Compress\&QA and Query Rewrite\&QA, which generate intermediate results, outperform single-model approaches. This highlights the potential of chain-of-model collaboration. Our FTHSS method further optimizes this by reducing redundant computations, yielding significant efficiency gains.

\paragraph{FTHSS outperforms Distill in both single-round and multi-round scenarios.} While Distill attempts to fine-tune a single model to handle all intermediate steps, distilling multiple models’ capabilities into one, this approach presents notable challenges. It requires the model to excel across all intermediate tasks; otherwise, the final result may be compromised. As shown in Table \ref{tab:recomp_performance}, \ref{tab:rewrite_performance}, \ref{tab:memory_performance}, and \ref{tab:rpg_performance}, experimental results reveal that distilling the capabilities of multiple models into a single model leads to varying degrees of performance degradation in both single-round and multi-round tasks. This underscores the superiority of FTHSS, where each model is allowed to specialize in its strengths, resulting in improved overall performance.

\subsection{Inference Efficiency Improvements}
To demonstrate the efficiency of our method, we present latency speed-ups achieved by eliminating redundant forward passes over intermediate results. We compare the inference latency of model B in FTHSS with that of the original model chain (where model A's output serves as input to model B), evaluating various intermediate result lengths. Results are averaged over 10 runs, performed on an Nvidia L20 GPU with the Llama-3-8B architecture.

Table~\ref{tab:latency} shows that for input sequences of 3,000 tokens, FTHSS reduces inference latency to less than one-third of the original model's. This improvement demonstrates that FTHSS maintains accuracy while significantly reducing latency. For sequences of 250 tokens, however, the speed-up is minimal due to GPUs' efficient parallel processing, limiting acceleration for smaller token counts.

In multi-round tasks, where each model in the chain may be repeatedly invoked, multiple copies of KV Caches are typically stored. FTHSS addresses this by enabling shared KV hidden states across models, reducing KV Cache storage to a single instance, regardless of chain length. As shown in Table \ref{tab:latency}, FTHSS significantly reduces GPU memory usage compared to a standard model chain. For the Llama-3-8B architecture, the KV Cache size for an input sequence of 1000 tokens is 137.5 MB. When multiple models are used, FTHSS saves $(n-1) \times 137.5 \text{ MB}$ of GPU memory. Thus, in multi-round tasks, FTHSS not only eliminates redundant computations, reducing latency, but also removes the need for multiple KV Cache copies, resulting in substantial memory savings.

\subsection{Further Analysis}
\begin{table}[t]
    \centering
    \resizebox{0.4\textwidth}{!}{
    \begin{tabular}{cccccc}
        \toprule
        \multicolumn{6}{c}{Inference latency(s)(single-round task)} \\
        \midrule
        \textbf{tokens} & Chain of models & FTHSS \\
        \midrule
        250 & $0.45$ & $0.41$ \\
        500 & $0.52$ & $0.42$ \\
        1000 & $0.66$ & $0.43$ \\
        3000 & $1.44$ & $0.46$ \\
        \midrule \midrule
        \multicolumn{6}{c}{KV cache size(MB)(multi-round task)} \\
        \midrule
        \textbf{Models} & Chain of models & FTHSS \\
        \midrule
        1 & $137.5$ & $137.5$ \\
        2 & $137.5*2$ & $137.5$ \\
        3 & $137.5*3$ & $137.5$ \\
        \bottomrule
    \end{tabular}
    }
\caption{\textbf{Top:} Inference latency of model B in the chain A $\rightarrow$ B, with varying intermediate result lengths (in tokens), while output length is fixed at 16.
\textbf{Bottom:} GPU memory occupancy for KV cache under varying model counts in multi-round tasks, with total length fixed at 1000. Latencies are measured on an NVIDIA L20, with KV states stored in bfloat16.}
    \label{tab:latency}
\end{table}

\begin{figure}[t]
\centering
\includegraphics[width=0.49\textwidth]{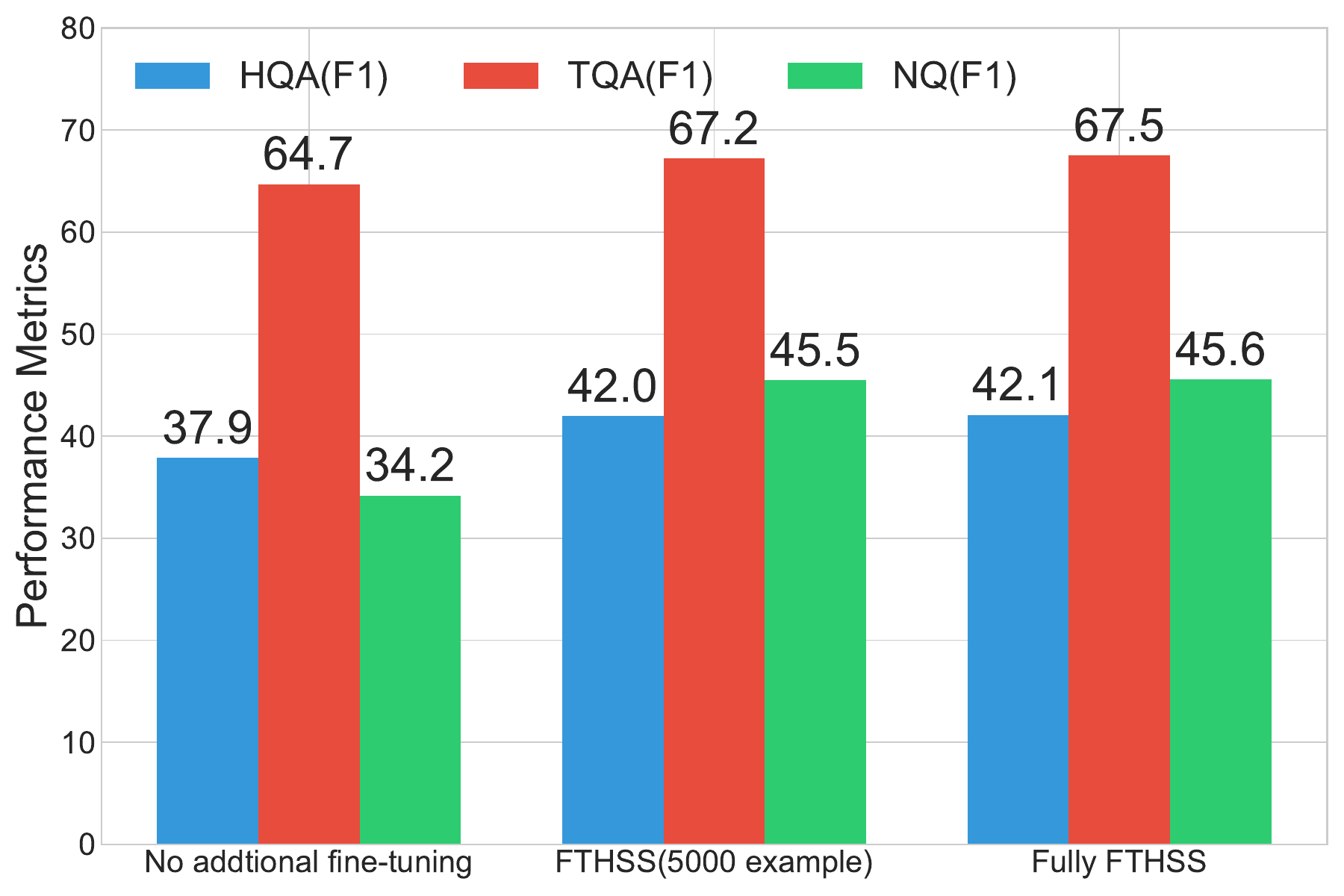}
\caption{Performance comparison of three fine-tuning strategies on Context Compression \& QA task: (1) No additional fine-tuning, using noisy KV hidden states directly; (2) FTHSS (5000 samples), where the standard-prompt-tuning model is fine-tuned on 5,000 examples; and (3) Fully FTHSS, where the base model undergoes full-dataset fine-tuning.}
\label{fig:training_scale}
\end{figure}
In practice, specialized models are already trained using methods like Prompt Tuning. To apply our approach and simplify the model chain, these models may require re-fine-tuning, which can be computationally expensive. This raises the question: can these trained models—trained on plain text instead of the KV hidden states of previous models—be used with minimal or no fine-tuning?

Figure \ref{fig:training_scale} compares three approaches: (1) the standard prompt-tuned model without additional re-fine-tuning (Standard), which attempts to interpret the noisy KV hidden states of the previous model directly; (2) continuing fine-tuning the standard prompt-tuned model on 5,000 examples using FTHSS (5000 samples); and (3) fully fine-tuning the base model on the entire dataset with FTHSS (Fully FTHSS). Experiments on the Context Compression\&QA task show that even the standard fine-tuned model generates mostly correct answers. On the TQA dataset, the F1 score of the standard model is close to that of the fully fine-tuned model using FTHSS. However, performance drops on more complex datasets like HotpotQA and NQ due to noise in the KV hidden states. Additionally, fine-tuning models on a small dataset significantly improves performance. This suggests that fine-tuning standard prompt-tuned models on a small dataset using FTHSS is sufficient to mitigate noise, making full-dataset re-fine-tuning unnecessary.



As discussed in Section \ref{sec:fine-tuning}, passing KV hidden states between models mainly introduces  unnecessary attention to noisy tokens. Previous work, such as attention sink~\cite{xiao2023efficient}, has shown that attention exhibits sparsity properties, meaning that a few noisy tokens do not significantly impact the final output. Consequently, using a standard prompt-tuned model without further fine-tuning can still yield strong performance on simpler tasks.

\section{Conclusion}
In this paper, we introduced FTHSS, a method that enables models in a chain to directly share KV hidden states, eliminating redundant forward passes over intermediate results and reducing KV cache storage. By reordering the input and attention masks at each layer, FTHSS allows downstream models to leverage KV hidden states from upstream models.  Our experiments demonstrate that FTHSS matches the performance of traditional model chains while significantly improving the inference efficiency in both single-round and multi-round scenarios. 


\section{Limitations}
While our method is effective for open-source models, it cannot be directly applied to closed-source models that only provide API access, limiting its applicability in such settings. Additionally, because the method involves fine-tuning, experiments were not conducted on large models, such as those with 70B parameters, due to computational resource constraints. Future work should explore the generalizability of hidden state sharing methods in larger models and across diverse, high-quality, and challenging datasets.

\bibliography{anthology}

\appendix
\begin{table}[t]
    \centering
    \begin{tabular}{cc}
        \toprule
        \textbf{Dataset name} & \textbf{Train/Test} \\
        \midrule
        \multicolumn{2}{c}{\textbf{Context Compression\&QA task}} \\
        Natural Questions(NQ) & 39,466/3,610 \\
        TriviaQA(TQA) & 47,531/11,313 \\
        HotpotQA(HQA) & 26,556/500 \\
        \midrule
        \multicolumn{2}{c}{\textbf{Query Rewrite\&QA task}} \\
        Training Data~\cite{ma2023query} & 37,520/-\\ 
        2WikiMultiHop(2Wiki) & -/500 \\
        HotpotQA(HQA) & -/500 \\
        \midrule
        \multicolumn{2}{c}{\textbf{Memory\&Reasoning task}} \\
        Training Data~\cite{jin2024disentangling} & 10,925/- \\
        StrategyQA & -/687 \\
        TruthfulQA(TruthQA) & -/164 \\
        CommonsenseQA(ComQA) & -/1,221 \\
        \midrule
        \multicolumn{2}{c}{\textbf{Active RAG task}} \\
       Training Data~\cite{lyu2024retrieve} & 47,689/- \\
        2WikiMultiHop(2Wiki) & -/500 \\
        Pubhealth & -/987 \\
        \bottomrule
    \end{tabular}
    \caption{Dataset statistics.}
    \label{tab:datasets}
\end{table}

\section{Hyperparameters and Datasets}
\label{sec:hyper}

\paragraph{Hyperparameters.} We fine-tune all parameters of our models for up to 3 epochs on 4 Nvidia A6000 GPUs. Our learning rate is 2e-4, and the gradient accumulation step is set to 8. We use
3\% of steps for linear warm-up of the learning rate and decay it linearly to 0 over training. To save
memory, we use DeepSpeed ZeRo-2~\cite{rajbhandari2020zero,rasley2020deepspeed} optimization, gradient checkpointing, and BF16 mixed precision training. During training, we use a maximum sequence length of 1224 for every sample, 100 learnable prompt tokens, and finetune using the Adam optimizer~\cite{kingma2014adam} with no weight decay. Our training script is based on HuggingFace accelerate~\cite{accelerate} libraries.

All base models in this paper are Llama-3-8B-Base unless otherwise specified. All PEFT fine-tuning methods are based on Prompt tuning, with the number of learnable prompt tokens set to 100. For methods that do not involve model training (e.g., Native, Standard RAG, and CoT), we utilize Llama-3-8B-Instruct, as its instruction-following capability is essential for these approaches.

\paragraph{Datasets.} The statistical details of the training and test datasets used in the experiments are provided in Table \ref{tab:datasets}. In the Context Compression\&QA task, the training phase utilizes the augmented NQ, TQA, and HQA datasets from recomp~\cite{xu2023recomp}. These datasets were created by using ChatGPT to semantically compress retrieved documents into concise summaries, generating synthetic training data. For model evaluation, we use the full test sets of NQ and TQA, along with a subset of the HQA development set, as validation benchmarks to ensure a comprehensive and reliable assessment of model performance. In the Query Rewrite\&QA task, we use the dataset from ~\citet{ma2023query} for training and evaluate the model on the multi-hop question datasets HQA and 2Wiki. In the Activate RAG task, we use the dataset from ~\citet{lyu2024retrieve} for training and evaluate the model on the short-form QA dataset PubHealth and the multi-hop QA dataset 2Wiki.

As for the retrieved documents, by default, we use the top one document ranked by Contriever-MS MARCO~\cite{izacard2021unsupervised} on Wikipedia corpus from Dec. 20, 2018, which is done to ensure a fair comparison among all baseline models. In the Query Rewrite\&QA and Active RAG tasks,  we use the top one document ranked by the BM25 retrieval algorithm. Improving the retriever is not the primary focus of this work; therefore, the retriever selection criterion is to maintain consistency with the papers that proposed these tasks.

\section{Task Explanation}
\label{sec:task}
\begin{table*}[t]
    \centering
    \resizebox{\textwidth}{!}{
    \begin{tabular}{p{17cm}}
        \hline
        \textbf{Input}: How many episodes are there in dragon ball z?(\textcolor{gray}{\textit{NQ}})\\
        \hline
        \textbf{Input of the compression model}: \textcolor{gray}{<retrieved>April 5, 2009, the series premiered in Japan airing in Fuji TV. \"Dragon Ball Z Kai\" reduced the episode count to 159 episodes (167 episodes internationally), from the original footage of 291 episodes. Damaged frames were removed, resulting in some minor shots being remade from scratch in order to fix cropping, and others to address continuity issues. The majority of the international versions, including Funimation Entertainment's English dub, are titled \"Dragon Ball Z Kai\". premiered on Fuji TV on February 7, 1996 and ran until November 19, 1997 for 64 episodes. Unlike the first two anime series, it is not. Edition,\" which collects three individual volumes into a single large volume. However, in 2013 Viz began publishing new 3-in-1 volumes collecting the entire manga series, including what they previously released as \"Dragon Ball Z\", under the \"Dragon Ball\" name. The \"Dragon Ball Z\" films comprise a total of 15 entries as of 2015...</retrieved>} \\
        \textbf{Output of the compression model}:  \textcolor[rgb]{1,0.65,0}{<compressed>P\"Dragon Ball Z Kai\" has a total of 159 episodes (167 internationally), reducing the original count of 291 episodes. The last aired episode of \"Dragon Ball Kai\" in Japan was the 97th episode, with the 98th episode being released direct-to-video on August 2, 2011.</compressed>} \\
        \textbf{Output of the question answering model}:  \textcolor[rgb]{0.9,0,0.1}{<final\_answer>291 episodes</final\_answer>}\\
        \hline
    \end{tabular}
    }
    \caption{An Example for NQ datasets in Compression\&QA task. Model chain: compression model$\rightarrow$question answering model.}
    \label{tab:nq data examples}
\end{table*}

\begin{table*}[t]
    \centering
    \resizebox{\textwidth}{!}{
    \begin{tabular}{p{17cm}}
        \hline
        \textbf{Input}: Which one was established first, Grouplogic or Inbios?(\textcolor{gray}{\textit{2Wiki}})\\
        \hline
        \textbf{Output of the query rewriting model}:  \textcolor[rgb]{1,0.65,0}{<query\_rewrite>Grouplogic establishment year; Inbios establishment year</query\_rewrite>} \\
        \textbf{Input of the question answering model}: \textcolor{gray}{<retrieved>GroupLogic, Inc., founded in 1988 and headquartered in Arlington, Virginia, USA, is an enterprise software company that develops, sells and supports software for moving and storing data including activEcho, mobilEcho, ArchiveConnect, MassTransit and ExtremeZ- IP...}
        \\
        \textcolor{gray}{InBios International, Inc. is a medical diagnostic company based in Seattle that specializes in the detection of infectious diseases. The company was founded in 1996, and since its inception has developed several technologies useful in designing rapid and ELISA based immunodiagnostic assays. In April 2011, InBios International announced the clearance of its IgM test for dengue fever, DENV Detect IgM Capture ELISA...</retrieved>} \\
        \textbf{Output of the question answering model}: \textcolor[rgb]{0.9,0,0.1}{<Final answer>291 episodes</Final answer>}\\
        \hline
            \end{tabular}
        }
    \caption{An Example for 2Wiki datasets in Query Rewriting\&QA task. Model chain: query rewriting model$\rightarrow$question answering model.}
    \label{tab:2wiki data examples}
\end{table*}

\begin{table*}[t]
    \centering
    \resizebox{\textwidth}{!}{
    \begin{tabular}{p{17cm}}
        \hline
        \textbf{Input}: Question: Sammy wanted to go to where the people were.  Where might he go? Options: A.race track B.populated areas C.the desert D.apartment E.roadblock
        (\textcolor{gray}{\textit{ComQA}})\\
        \hline
        \textbf{Output}: \textcolor[rgb]{0.9,0,0.1}{<reason\_start>Let's reason with knowledge: The objective is to determine a location where there are many people, as per Sammy's desire to be where the people are.  Each option needs to be evaluated based on the likelihood of it being a place with many people.<reason\_end>}\textcolor[rgb]{1,0.65,0}{<memory\_start>Let's recall the knowledge: A race track can be very crowded during events. However, it may not always be populated, so it lacks consistency in being populated.  Populated areas are locations where there are a high number of people living or congregating, such as cities and towns. This option aligns closely with the goal of being where people are consistently present.  Deserts are typically sparsely populated because of harsh living conditions, making them unlikely places for finding many people.  While an apartment building may contain a number of residents, it is not as widely or consistently populated as broader populated areas like cities.  A roadblock is usually not a gathering place for people. It might temporarily have a group or line of vehicles, but it is not known for large groups of people congregating for extended periods.<memory\_end>}\textcolor[rgb]{0.9,0,0.1}{<reason\_start>Let's reason with knowledge: Based on the assessments, option B (populated areas) is the most appropriate choice as it consistently fulfills the objective of being a place where there is a high number of people present. [The answer is]: B <reason\_end>}\\       
        \hline
            \end{tabular}
        }
    \caption{An Example for CommonsenceQA datasets in Memory\&Reasoning task. The reasoning and memory models are iteratively called multiple times during inference.}
    \label{tab:comqa data examples}
\end{table*}

\begin{table*}[t]
    \centering
    \resizebox{\textwidth}{!}{
    \begin{tabular}{p{17cm}}
        \hline
        \textbf{Input}: Which movie came out first Escape to Witch Mountain or Pete's Dragon?(\textcolor{gray}{\textit{HotpotQA}})\\
        \hline
        \textbf{Output}: \textcolor[rgb]{0.9,0,0.1}{<plan\_start>Escape to Witch Mountain release<plan\_end>}\textcolor{gray}{<retrieved>Escape to Witch Mountain is a 1975 American fantasy-children's film, adapted from the 1968 science fiction novel of the same name written by Alexander H. Key</retrieved>}\textcolor[rgb]{1,0.65,0}{<answer\_start>Escape to Witch Mountain came out first,<answer\_end>}\textcolor[rgb]{0.9,0,0.1}{<plan\_start>Pete's Dragon release<plan\_end>}\textcolor{gray}{<retrieved>Pete's Dragon is a 2016 American fantasy comedy-drama adventure film directed by David Lowery, written by Lowery and Toby Halbrooks, and produced by James Whitaker. </retrieved>}\textcolor[rgb]{1,0.65,0}{<answer\_start>before Pete's Dragon. <answer\_end>}[Combine]\textcolor[rgb]{1,0.65,0}{<answer\_start>Escape to Witch Mountain<answer\_end>}\\
        \hline
            \end{tabular}
        }
    \caption{An Example for HotpotQA datasets in Active RAG task.}
    \label{tab:hotpotqa data examples}
\end{table*}

In this section, we provide detailed examples to demonstrate why the evaluation tasks used in this paper involve multiple models. Table~\ref{tab:nq data examples} illustrates the Compression\&QA task. The documents retrieved by RAG are often excessively long and contain a significant amount of noise, which can mislead the question-answering model if input directly. By first using a model to compress the documents and then providing its output as input to the question-answering model, the accuracy of the responses can be significantly improved. The model chain in the Compression\&QA task is designed based on this approach, consisting of a summarization model whose output serves as the input to the question-answering model.

Table ~\ref{tab:2wiki data examples} presents the Query Rewriting\&QA task. For complex problems such as multi-hop QA, directly using the question as a query often fails to retrieve the appropriate context. To address this, we utilize another model to rewrite the query, which is then used to retrieve more accurate contextual information, followed by inputting this refined context into the question-answering model. The Query Rewriting\&QA task also involves a model chain.

Tables~\ref{tab:comqa data examples} and ~\ref{tab:hotpotqa data examples} demonstrate the model chains in multi-round scenarios, where the models are not invoked only once but are iteratively called. In the Memory\&Reasoning task, the model first recalls the knowledge required to answer the question, then uses this recalled knowledge to reason and generate the answer. Since these two sub-tasks differ significantly, different models must be deployed to handle them separately. Furthermore, a single round is insufficient to ensure that all required knowledge is retrieved, so these two sub-tasks need to be executed alternately and repeatedly. Additionally, the Active RAG task involves multiple rounds of retrieval, where the model dynamically decides what to retrieve during the generation process (the planning phase), followed by generating the response based on the retrieved information (the answering phase). The planning and answering sub-tasks are iteratively performed, requiring two distinct models to be deployed.

\section{Inference Details}

Algorithm \ref{alg:multi_prompt_inference} illustrates the process of inference in a single-round task, where multiple prompt-tuning-based models share KV hidden states. The performance of a shared base model across different tasks depends on the learnable, task-specific prompt tokens. During inference, these prompt tokens are dynamically switched, as demonstrated in line 12 of the algorithm. Furthermore, sharing hidden states implies that the KV cache from the previous model can be reused directly, without the need to recompute the intermediate KV hidden states of \( Y_i \). The red-striped portion in the algorithm shows the computational savings of our approach compared to previous prompt-tuning methods.

Assuming the total length of intermediate results is $n$, the computational savings of this algorithm are $O(n^2)$, given the quadratic complexity of the transformer.

Notably, the computational savings occur during the prefilling phase, which runs in parallel. Therefore, when the length of the intermediate results is relatively short, the savings have a minimal impact on the inference latency.

\begin{algorithm*}[t]
\caption{The Inference Process of FTHSS in Single-Round Tasks(The red-striped portion represents operations that are necessary for the original model chain, but are optimized and removed in FTHSS).}
\label{alg:multi_prompt_inference}
\begin{algorithmic}[1]
\STATE \textbf{Input:} Input sequence $X = (x_1, x_2, \dots, x_n)$
\STATE \textbf{Output:} Sub-task output sequences $Y_1 = (y_{11}, y_{12}, \dots)$, $Y_2$, $\dots$, $Y_t$

\STATE \textbf{Initialize:} 
\STATE \quad - Decoder-only transformer $T$ with parameters $\theta$
\STATE \quad - Task-specific soft prompt tokens $\{P_1, P_2, \dots, P_t\}$
\STATE \quad - KV Cache: $\text{Cache} \gets T(X)$ \quad (Encode input sequence)
\STATE \quad - Intermediate results: $Y_0$

\FOR{$i = 1$ \TO $t$}
\STATE \textbf{Prefilling Phase:}
\STATE \textcolor{red}{\sout{$\text{Cache} \gets \emptyset$}}
\STATE \textcolor{red}{\sout{$\text{Cache} \gets T(P_i, Y_{i-1}, \text{Cache})$}}
\STATE $\text{Cache} \gets T(P_i, \text{Cache})$ \quad (Compute and cache task-specific KV)
\STATE Initialize output sequence: $Y_i \gets [\text{\textless start\textgreater}]$

\STATE \textbf{Decoding Phase (Autoregressive):}
\FOR{$k = 1$ \TO max\_length} 
\STATE \quad 1. Current token: $y_{k-1} \gets Y_i[-1]$ \quad (Last generated token)
\STATE \quad 2. Compute embedding: $e_k \gets E(y_{k-1})$
\STATE \quad 3. Update decoder layers with KV Cache:
\STATE \quad \quad $h_k, \text{Cache} \gets T(e_k, \text{Cache})$ \quad (Reuse cached KV)
\STATE \quad 4. Compute logits: $p(y_k) \gets \text{softmax}(W_o h_k)$ \quad ($h_k$ from last layer)
\STATE \quad 5. Sample next token: $y_k \sim p(y_k)$
\STATE \quad 6. Append $y_k$ to $Y_i$
\ENDFOR
\ENDFOR

\STATE \textbf{Return:} Output sequences $Y_1, Y_2, \dots, Y_t$
\end{algorithmic}
\end{algorithm*}

\section{Position ID Rearrangement}
\label{sec:position}
If  $l$ is the last position ID of the preceding model, the position encoding of the current model should begin at $l+1$, ensuring that the accuracy of the attention computation during inference is unaffected. This is because, under Rotary Position Embedding (RoPE), the position ID ranges $[0, 1, \dots, l]$ and $[l+1, l+2, \dots, 2l+1]$ are equivalent in attention computation. The proof of this conclusion is presented below.

We prove that RoPE computes attention based solely on the relative position \( m-n \) , independent of the absolute positions \( m \) or \( n \). Given a query vector \( \boldsymbol{q}_m \) at position \( m \) and a key vector \( \boldsymbol{k}_n \) at position \( n \), RoPE applies rotations:

\begin{equation}
\begin{aligned}
\boldsymbol{q}_m & = R_m \boldsymbol{q}, \quad R_m = \begin{bmatrix}
\cos m\theta & -\sin m\theta \\
\sin m\theta & \cos m\theta
\end{bmatrix}, \\
\boldsymbol{k}_n & = R_n \boldsymbol{k}, \quad R_n = \begin{bmatrix}
\cos n\theta & -\sin n\theta \\
\sin n\theta & \cos n\theta
\end{bmatrix},
\end{aligned}
\end{equation}

where \( \theta \) is a frequency parameter. The attention score is:

\begin{equation}
\text{Score}(m, n) = \boldsymbol{q}_m^\top \boldsymbol{k}_n = (\boldsymbol{q}^\top R_m^\top)(R_n \boldsymbol{k}).
\end{equation}

Since rotation matrices are orthogonal (\( R^\top R = I \)), and satisfy \( R_m^\top R_n = R_{n - m} \), the score can simplify to:

\begin{equation}
\text{Score}(m, n) = \boldsymbol{q}^\top R_{n - m} \boldsymbol{k},
\end{equation}
which depends only on \( (n - m) \). For high-dimensional vectors, RoPE divides the vector into \( d/2 \) subspaces, applying rotations independently in each subspace:

\begin{equation}
R_{m}^{(i)} = \begin{bmatrix}
\cos m\theta_i & -\sin m\theta_i \\
\sin m\theta_i & \cos m\theta_i
\end{bmatrix},
\end{equation}

yielding:

\begin{equation}
\text{Score}(m, n) = \sum_{i=1}^{d/2} \boldsymbol{q}_i^\top R_{n - m}^{(i)} \boldsymbol{k}_i.
\end{equation}

Thus, RoPE strictly encodes relative positions, eliminating absolute position dependence. This property 
has been utilized in some precomputed KV cache scenarios~\cite{lu2024turborag}.

\end{document}